\documentclass{article}

    \PassOptionsToPackage{numbers}{natbib}

\usepackage[preprint]{neurips_2020}

\usepackage{graphicx} 
\usepackage{float} 
\usepackage[utf8]{inputenc} 
\usepackage[T1]{fontenc}    
\usepackage{xcolor}
\usepackage{color}
\usepackage{hyperref}       
\hypersetup{
    colorlinks=true
}
\usepackage{array}
\usepackage{url}            
\usepackage{booktabs}       
\usepackage{amsmath,amsfonts}       
\usepackage{nicefrac}       
\usepackage{microtype}      
\usepackage{algorithmic}
\usepackage{enumitem}

\title{Rethinking Distributional Matching Based \\Domain Adaptation}

\author{%
  Bo Li$^{1}\footnotemark[3]$ , Yezhen Wang$^{23}\footnotemark[3]$ , Tong Che$^{4}\footnotemark[3] ~ \footnotemark[1]$ , Shanghang Zhang$^{1}\footnotemark[3]$ ,\\
  \textbf{Sicheng Zhao}$^{1}$, \textbf{Pengfei Xu}$^{2}$, \textbf{Wei Zhou}, \textbf{Yoshua Bengio}$^{45}$, \textbf{Kurt Keutzer}$^{1}$\\
  $^{1}$UC, Berkeley $^{2}$DiDi, $^{3}$UC, San Diego \\
  $^{4}$MILA, Université de Montréal, $^{5}$CIFAR\\
}

\begin{document}
\maketitle
\renewcommand{\thefootnote}{\fnsymbol{footnote}} 
\footnotetext[3]{These authors contributed equally to this work. The order is determined by dice rolling.} 
\footnotetext[1]{Correspondence author.} 

\begin{abstract}
Domain adaptation (DA) is a technique that transfers predictive models trained on a labeled source domain to an unlabeled target domain, with the core difficulty of resolving distributional shift between domains. Currently, most popular DA algorithms are based on distributional matching (DM). However in practice, realistic domain shifts (RDS) may violate their basic assumptions and as a result these methods will fail. In this paper, in order to devise robust DA algorithms, we first systematically analyze the limitations of DM based methods, and then build new benchmarks with more realistic domain shifts to evaluate the well-accepted DM methods. We further propose InstaPBM, a novel Instance-based Predictive Behavior Matching method for robust DA. Extensive experiments on both conventional and RDS benchmarks demonstrate both the limitations of DM methods and the efficacy of InstaPBM: Compared with the best baselines, InstaPBM improves the classification accuracy respectively by $4.5\%$, $3.9\%$ on Digits5, VisDA2017, and $2.2\%$, $2.9\%$, $3.6\%$ on DomainNet-LDS, DomainNet-ILDS, ID-TwO. We hope our intuitive yet effective method will serve as a useful new direction and increase the robustness of DA in real scenarios. Code will be available at anonymous link: \href{https://github.com/pikachusocute/InstaPBM-RobustDA}{InstaPBM-RobustDA}.
\end{abstract}

\section{Introduction}
Recent deep learning algorithms suffer from a fundamental problem: models require large-scale labeled training data and fail to generalize well to unlabeled new domains. 
With deep neural networks' capability to learn rich representations, recent domain adaptation methods attempt to solve this problem by matching the source and target distributions in representation space~\cite{ajakan2014domain, adel2017unsupervised, ganin2016domain, becker2013non, ghifary2015domain, glorot2011domain, pei2018multi, long2015learning, jhuo2012robust,  Hoffman_cycada2017, zhao2019multi}. However, distributional matching (DM) cannot guarantee good generalization on the target domain and can even degrade results under realistic shifts~\cite{zhao2019learning, wu2019domain}. Some recent works \cite{long2018conditional, zhao2019learning, wu2019domain, combes2020domain} propose modified versions of distributional matching that resolve label distribution shift~\cite{wu2019domain} between domains. Although these alternatives can deal with some more general domain shifts, the domain shifts in the real world are still more complicated than commonly assumed, for we cannot restrict the variance, confounding factors, or peculiarities \cite{torralba2011unbiased, sturm2014simple} in target domain.
Here we would like to take a step back and rethink the DM methods: is it always helpful for domain adaptation to match source and target distributions?

Following this motivation, we first discuss the limitations of DM algorithms under realistic domain shifts (RDS) in Sec.~\ref{section:Limitations_of_DM}. For quantitatively analysis, we propose three representative RDS benchmarks in Sec.~\ref{section:RDS} and provide extensive empirical evaluations in Sec.~\ref{section:Experiments} to support our analysis. Given the observation that DM methods are not robust in tackling RDS, we rethink UDA from first principles and investigate the per-instance predictive behavior across domains. We believe a robust DA algorithm should be designed to deal with complexity and diversity in the real world. To promote robust DA algorithms on realistic domain shifts, we argue that one should not rely on DM methods whose performance are heavily affected by certain assumptions. Instead of matching distributions, we should directly match certain functional properties of a good predictor $p_\theta(y|x)$ that holds both in the source and target domains. We call the set of these properties "preditive behaviors". Since these predictive behaviors are about properties of the discriminative model $p_\theta(y|x)$, they add minimal assumptions to the distributions $p_S(x)$ and $p_T(x)$, leading to robustness in most cases. Following this idea, we propose a novel method: Instance-based Predictive Behavior Matching (InstaPBM) in Sec.\ref{section:Methods}. Overall, InstaPBM consists of: (1) Mutual Information Predictive Behavior Matching; (2) Contrastive Predictive Behavior Matching; (3) Mix-up Predictive Behavior Matching. Each part corresponds to one predictive behavior we identify for a good predictive model on both domains.
 
The contributions of this paper are summarized as follows: 
\begin{itemize}[leftmargin=*]
\setlength\itemsep{0em}
     \item We systematically analyze limitations of DM methods when tackling realistic domain shifts. 
     \item We propose RDS benchmarks and extensively evaluate DM methods to verify their limitations.
     \item We further propose InstaPBM to contrastively match predictive behavior across domains and achieve robust DA under RDS. 
     \item Extensive experiments on both conventional and RDS benchmarks demonstrate the efficacy of InstaPBM, which consistently outperforms the state-of-the-art methods by large margin: Compared with the best baselines, InstaPBM improves the classification accuracy respectively by $4.5\%$, $3.9\%$ on Digits5, VisDA2017, and $2.2\%$, $2.9\%$, $3.6\%$ on DomainNet-LDS, DomainNet-ILDS, ID-TwO. On RDS benchmarks, InstaPBM shows stronger robustness to realistic shifts compared with DM methods.
\end{itemize}
%
\section{Limitations of Distributional Matching based DA}
\label{section:Limitations_of_DM}

\begin{figure*}
    \centering
    \includegraphics[width=0.95\linewidth]{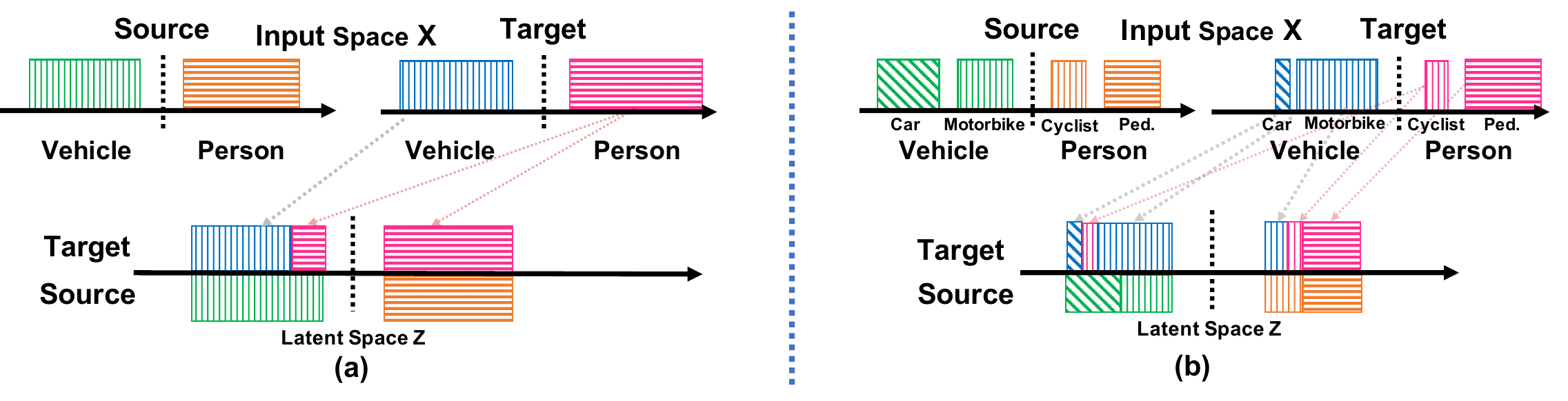}
    \caption{Failure cases of DM methods RDS. Colors denote categories, fill patterns denote sub-types, ped. denotes \textit{pedestrians}.}
    \label{fig:FC}
\end{figure*}

\textbf{Notations and Setup.}
In this paper we focus on the unsupervised domain adaptation (UDA) and take classification problem as a running example. We use $\mathcal{X}$ and $\mathcal{Y}$ to denote the input and output space respectively, and $\mathcal{Z}$ for the representation space generated from $\mathcal{X}$ by a feature transformation $g: \mathcal{X} \mapsto \mathcal{Z}$. Following, we denote a labeling function $h: \mathcal{Z} \mapsto \mathcal{Y}$ and a composite predictive transformation $f = g \circ h: \mathcal{X} \mapsto \mathcal{Y}$. Specifically, $f, g, h$ are parametrized by $\theta, \phi, \psi$, respectively. We use $X, Y, Z$ as random variables from space $\mathcal{X}, \mathcal{Y}, \mathcal{Z}$, respectively. In the setting of unsupervised domain adaptation, we have access to labeled source data $(x, y) \sim \mathcal{P}_S(X, Y)$ and unlabeled target data $x \sim \mathcal{P}_T(X)$, where $S$ and $T$ denote source and target domain.

Distributional matching (DM) is a core component in most recent UDA methods where they use either domain adversarial classifier \cite{tzeng2017adversarial, long2018conditional, hong2018conditional, he2019multi, xie2019multi, zhu2019adapting, zheng2020cross} or discrepancy-based approaches \cite{long2015learning, lee2019sliced, roy2019unsupervised, chen2019homm}.
In short, the purpose of distributional matching is to learn a shared feature representation space $\mathcal{Z}$ and a feature transformation $g: \mathcal{X} \rightarrow \mathcal{Z}$ such that $\mathcal{P}_u(Y|X)=\mathcal{P}_u(Y|Z=g(X)), u\in \{S,T\}$. To achieve domain adaptation, many algorithms focus on minimizing the source domain errors and the distance between source and target distributions in the latent space:
\begin{equation}
   \mathop{\text{min}} \limits_{\phi, \psi} L_S(\phi, \psi) + D(\mathcal{P}_S(Z) \ \ || \ \ \mathcal{P}_T(Z)) + \Omega(\phi, \psi)
   \label{eq:dann_objective}
\end{equation}
where $\phi$ and $\psi$ denote the parameters of $g$ and $h$, $D$ is a distance metric of distributions, $Z = g(X)$, and $\Omega$ is a regularizer term. Conventional instantiations of $D$ can be Maximum Mean Discrepancy~\cite{huang2007correcting}, domain classifier with Jensen-Shannon (JS) divergence~\cite{ganin2016domain} or Wasserstein distance~\cite{shen2018wasserstein}. 
Other generative or translation based DA \cite{Hoffman_cycada2017, zhao2019multi} can also be included in this framework as they match distribution $\mathcal{P}_T(X)$ and $\mathcal{P}_S(X)$ over pixel space $\mathcal{X}$ instead of representation space $\mathcal{Z}$. To this end, we can induce the most common hypothesis of DM: $\mathcal{P}_S(Z)=\mathcal{P}_T(Z), Z=g(X)$.

First, we prove that if the distributions of source domain and target domain are disjointed (which is most common case for high dimensional data such as images), then there is no guarantee for target label correctness even with a perfect distributional matching of representations. Specifically, for any arbitrary wrong label matching, there is a distributional matching which perfectly matches the representation, but wrongly matches labels in a way specified by the wrong label matching. This result tells us matching distributions is not sufficient for correctly matching labels between two domains. We detailed this into supplementary material due to space limitation.

Besides this theoretical results, exact matching distributions over representation space $\mathcal{Z}$ will result in incorrect matches when $\mathcal{P}_S(Y) \neq \mathcal{P}_T(Y)$ and lead to unbounded target errors. We call this case label distributional shift (LDS \cite{wu2019domain,combes2020domain}). Consider the case of binary classification with non-overlapping support. Suppose the source contains $50\%$ vehicle and $50\%$ person, while the target contains $70\%$ vehicle and $30\%$ person. Successfully aligning these distributions in representation space requires the classifier to predict the same fraction of vehicle and person on source and target. If one achieves $100\%$ accuracy on the source, then target accuracy will be at most $80\%$ as illustrated in Fig. \ref{fig:FC}(a). 
Our experimental results in Sec.~\ref{section:Experiments} also demonstrate this issue under LDS.

To tackle label distribution mismatch, another DM hypothesis: $\mathcal{P}_S(Z|Y = y)=\mathcal{P}_T(Z|Y = y), \forall y \in \mathcal{Y}$ is proposed alongside \cite{combes2020domain}. However, such hypothesis may fail in the presence of intermediate layer distribution shift (ILDS), which indicates the intra-class feature distribution $P(Z|Y)$ may be different in two domains. A special case of ILDS is the sub-label distribution shift, where the distributions of sub-classes are different between source and target. Take the classification between \textit{vehicle} and \textit{person} as an example. Assume the source \textit{vehicle} class contains $50\%$ \textit{car} and $50\%$ \textit{motorcycle}, while the target \textit{vehicle} class contains $10\%$ \textit{car} and $90\%$ \textit{motorcycle}. Aligned with our analysis for LDS, since there are much higher proportion of \textit{motorcycle} class in target, a prefect conditional classifier trained on uniformed source domain will be biased to the source sub-label distribution and mis-predict the extra amount of target \textit{motorcycle} to another category: \textit{person}, as illustrated in Fig. \ref{fig:FC}(b).
Such phenomenon usually happens when the divergence between sub-classes (intra-class gap) is ``larger'' than the inter-class gap.

Another recent DM method~\cite{wu2019domain} attempts to address LDS by asymmetrically relaxing the distribution alignment with hypothesis: $\frac{\mathcal{P}_T(Z)}{\mathcal{P}_S(Z)} \leq 1 + \beta \ \ \text{for some} \ \ \beta > 0$. Such method proposes a relaxed distance for aligning data across domains that can be minimized without requiring latent-space distributions to match exactly. Even though the asymmetrical relaxation of distribution alignment mitigates the limitations of domain-adversarial algorithms~\cite{tzeng2017adversarial, long2018conditional} under LDS, it is not sufficient for justification and suffers from the following issues: (1) To learn correct matching, one may have to select large $\beta$, making the objective itself too loose to learn discriminative target representations. (2) There may be some samples from target lying outside the support of source in the latent space, in which case the density ratio $\frac{\mathcal{P}_T(Z)}{\mathcal{P}_S(Z)}$ is unbounded, and the upper bound of target risk becomes vacuous. We name this case Target with Outliers(TwO). (3) Although the perfect alignments exist, there may also be other alignments that satisfy the objective but predict poorly on the target data.

In general, these hypotheses demonstrate the purpose of most DM methods' theoretical study and algorithms. While they can make the performance guarantees vacuous under RDS, thus, we conclude that to achieve robust DA, we must devise DA algorithms without assuming any previously mentioned assumptions. We show how to devise such an algorithm in Sec.~\ref{section:Methods}.

\section{Realistic Domain Shift Benchmarks}
\label{section:RDS}

In order to quantitatively verify the analysis of realistic domain shifts which limit the performance of DM methods in Sec.~\ref{section:Limitations_of_DM}, we propose new DA benchmarks with three types of RDS, which better reflect the real world scenarios and can promote UDA research with more practical effectiveness:

\textbf{Label Distribution Shift (LDS).}
LDS indicates the mismatch between source and target label distributions. We insert LDS on public datasets Digits5~\cite{lecun1998gradient} and DomainNet~\cite{peng2019moment} by sampling target domain into long-tailed distribution among all classes (source domain unchanged). In this case, we do not require the source and target domains share the same visual looking in the same class, which is more aligned with real-world scenarios. 

\textbf{Intermediate Layer Distribution Shift (ILDS).}
ILDS indicates the intra-class feature distributions $P(Z|Y)$ are different in two domains, whose special case is sub-label distribution shift. We create ILDS on a public DA dataset DomainNet~\cite{peng2019moment} by combining the similar sub-classes into corresponding meta-class and adjusting the number of samples for each sub-class to form long-tailed distribution within each meta-class on the target (source domain unchanged).
 
\textbf{Target with Outliers (TwO).}
TwO indicates the target domain contains some samples that lie outside the support of source domain in the latent space $\mathcal{Z}$. To build such benchmark, we choose ImageNet as source and DomainNet as target, and add noisy data samples into the target.

Due to space limitation, we will provide more details about label, sub-label distributions on each dataset and examples of TwO in the supplementary material.

\section{Instance-based Predictive Behavior Matching}
\label{section:Methods}

Based on the analysis in Sec.~\ref{section:Limitations_of_DM} and Sec.~\ref{section:RDS}, DM methods are susceptible to more realistic domain shifts. To overcome these limitations, we propose an Instance-based Predictive Behavior Matching (InstaPBM) method to match the instance predictive behaviors (PBs) across domains. PBs are about the intrinsic properties of the predictive model $p_{\theta}(y|x)$ which are independent of small perturbations of underlying source and target distributions, so they does not rely on strong assumptions of underlying distributions, this is why training domain-adapted classifier by matching PBs can achieve robustness. For consistent PB matching, we first propose to utilize mutual information maximization (MIM) on target data to simulate the mutual information change in the source domain. We further propose contrastive PB matching and mix-up PB to match different kinds of predictive patterns on both domains. To this end, we design a set of operations $\mathcal{T}$, which can be divided into three sets: (1) Semantic Preserving Operations $\mathcal{T}_{sp}$. (2) Mix-up Interpolation Operations $\mathcal{T}_{\beta}$. (3) Semantic Transforming Operations $\mathcal{T}_{st}$. (see Sec. \ref{sub:Contrastive Predictive Behavior Matching}, \ref{sub:mixup predictive behavior matching} and \ref{sub:Auxiliary Semantic Information Matching}).  Other predictive behaviors matching may also be proposed, which we leave for future work. 

\subsection{Mutual Information Predictive Behavior Matching}
\label{subsection:Predictive Behavior on Target}
Learning algorithms often construct predictors with supervised information. During training, the mutual information $I(X; Y)$ of source data increases progressively, because the back-propagated supervised information leads to a strong correlation between the predicted label and input data. We opine that the target domain should also obey this pattern (or predictive behavior). However, a challenge emerges since the target data is unlabeled. To this end, we propose to use mutual information maximization to simulate this behavior on target domain, which can be equivalently expressed as: $I_{T}(X; Y) = H_{T}(Y) - H_{T}(Y | X)$.

Therefore the objective of mutual information $I_{T}(X;Y)$ maximization is equivalent to two parts: maximizing the info-entropy $H_{T}(Y)$ and minimizing the conditional entropy $H_{T}(Y|X)$. Intuitively, by minimizing $H_{T}(Y|X)$ on unlabeled data, we increase the discriminability of the predictor towards target domain, which drives the decision boundary away from the target data~\cite{NIPS2004_2740}. However, when the groundtruth label distribution is imbalanced, minimizing $H_{T}(Y|X)$ solely can be trivially done by predicting a 100\% probability on one particular class. Thus we need to maximize the $H_{T}(Y)$ in order to prevent output to collapse to a particular class, namely, ensure the diversity of the output~\cite{cui2020nnm}. It is worth noting that we do not expect $H_{T}(Y)$ to be as large as possible, instead, we maximize $H_{T}(Y)$ only when it is lower than a pre-defined threshold. The corresponding loss function is as follows:
\begin{equation}
    \mathcal{L}_M (T; \theta) = \mathbb{E}_{y\sim \mathcal{P}_{\theta}(Y)} [ \log p_{\theta}(y)] - \mathbb{E}_{x \sim \mathcal{P}_{T}(X)} [\sum_y p_{\theta}(y|x) \log p_{\theta}(y|x)]
\end{equation}
where $p_{\theta}(y|x)$ denotes probability of each sample output from predictor,  $\mathcal{P}_{\theta} (Y)$ represents the distribution of predicted target label $Y$. We use a moving average $q(y)$ of $p_\theta(y|x),x\sim \mathcal{P}_T(x)$ as an approximation of $p_\theta(y)$. Then we estimate the gradient of $H_\theta(Y)=\mathbb{E}_{y\sim \mathcal{P}_{\theta}(Y)} [ \log p_{\theta}(y)]$ as: 
\begin{equation}
    \tilde{\nabla} H_\theta(Y) = \sum_{(x,y)\sim \mathcal{P}_S} \nabla_\theta p_\theta(y|x) \log q(y) 
\end{equation}

\begin{figure}
    \centering
    \includegraphics[width=0.95\linewidth]{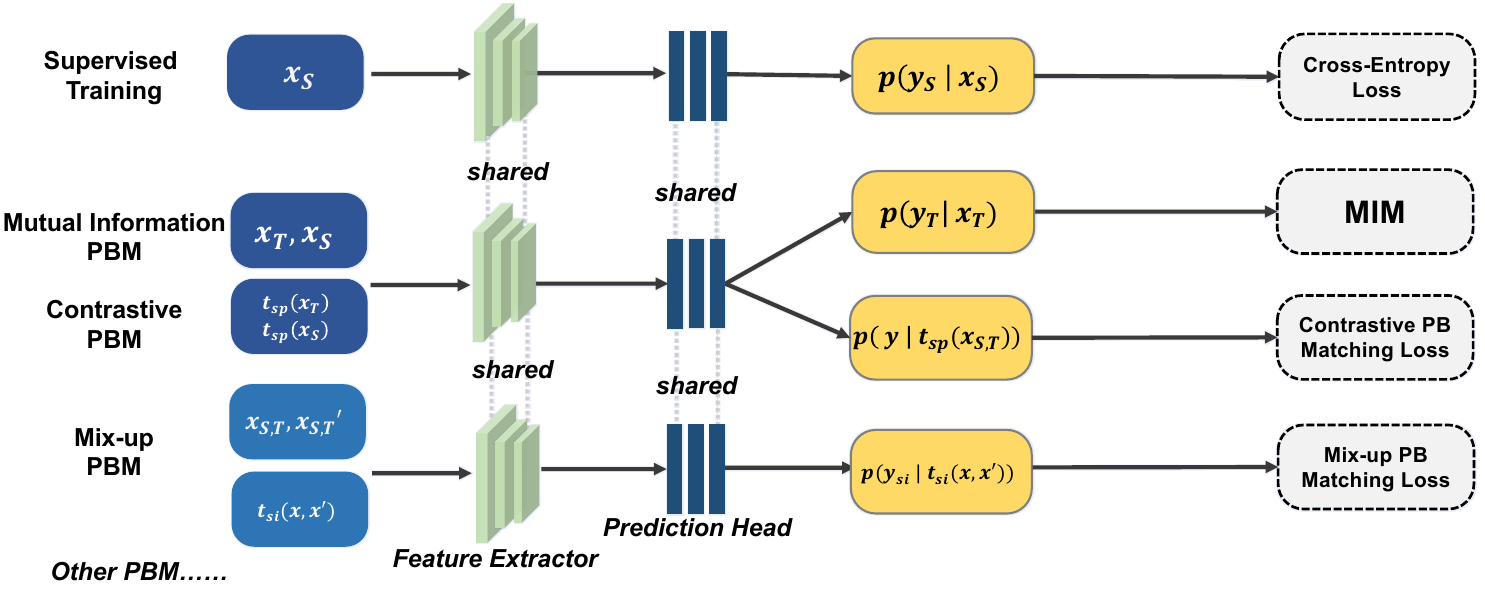}
    \caption{The proposed method of our InstaPBM, where $x_u \sim \mathcal{P}_u(X), u \in \{S, T\}$ denote data from source and target domain, $t \sim \mathcal{T}_{sp}$ and $t \sim \mathcal{T}_{st}$ denote semantic preserving and semantic transforming operations respectively, $y_{st}$ denote semantic transforming labels in self-supervise training.}
    \label{fig:pred_behavior_match}
\end{figure}

\subsection{Contrastive Predictive Behavior Matching}
\label{sub:Contrastive Predictive Behavior Matching}
Another predictive behavior that should be satisfied on both domains is about "augmentation consistency". Consider $t$ is an augmentation operation which preserves the semantics of the image $X$ (e.g. color distortion), then $t$ should not change the label of the image $t(X)$ after the operation. Here we introduce semantic preserving operations for CPBM, which will keep semantics of the instances. The predictive behavior we are matching using this loss can be described as "a good predictor on both domains should predict the same label for $X$ and $t(X)$, while (with a high probability) predict different labels for $X$ and $X'$ when $X$ and $X'$ belong to different classes". 

The semantic preserving $\mathcal{T}_{sp}$ operations on instances include two types:(1) Random Augmentation~\cite{xie2019unsupervised}; (2) Noise Injection~\cite{miyato2018virtual}. 
Following this idea, we propose an contrastive training loss to match the PB for instances before and after operations, and maximize the disagreement for instances from different classes. The model is jointly trained on source and target for the first objective and on the source for the second.
Formally, the objective can be written as 
\begin{equation}
\begin{aligned}
    \mathcal{L}_C (S,T; \theta) &= \mathbb{E}_{x \sim \mathcal{P}_{S, T}, t \sim \tau}[ D_{KL}(p_{\theta}(y \ | \ x) \ || \ p_{\theta}(y \ | \ t(x))) ]  \\ &- \lambda_{con} \mathbb{E}_{(x, y) \sim \mathcal{P}_{S}, (x', y') \sim \mathcal{P}_{S}, \ y \neq y'} [D_{KL} (p_{\theta}(y \ | \ x)\ || \ p_{\theta}(y' \ | \ x')) ]
    \label{eq:pred_bhav_matching}
\end{aligned}
\end{equation}

where $x'$ and $x$ belong to different classes. The training process can be divided into: (1) Apply a $t \sim \mathcal{T}_{sp}$ operation on input instance $x \sim \mathcal{P}_{S, T}$ from both domains to have the transformed version $t(x)$. Then feed $(x$, $t(x))$ into $p_{\theta}$ to compute $p_{\theta}(y \ | \ x)$ and $p_{\theta}(y \ | \ t(x))$. (2) Minimize KL divergence between the two distributions $D_{KL} (p_{\theta}(y | x)\ || \ p_{\theta}(y | x'))$. These operations advance the model with realistic augmented examples and create the same training objective of semantics maintaining~\cite{xie2019unsupervised}, so that this predictive behavior is consistently strengthen across domains.

\subsection{Mix-up Predictive Behavior Matching}
\label{sub:mixup predictive behavior matching}
Aligned with our predictive behavior matching objective, inspired by~\cite{zhang2017mixup}, we introduce Mix-up Predictive Behavior Matching (MuPBM) with linear interpolation operations $t(x, x', \beta) = \beta x + (1 - \beta) x' \sim \mathcal{T}_{\beta}, \beta \in [0,1]$ where $x, x'$ are raw input vectors sampled from both domains, and we create labels of $t(x, x', \beta)$ as $y_{si} = \beta y + (1 - \beta) y', \beta \in [0,1]$, where $y, y'$ are one-hot label encodings in source domain and softmax probabilities output from predictor in target domain. The predictive behavior should remain identical on both domains with introduced operations $\mathcal{T}_{\beta}$. Namely, MuPBM ensures a smooth transition between classes and encourages the model to behave similarly on linear interpolations of samples in both domains. Our objective enforces consistency and reduce undesirable oscillations by jointly training on both domains to deal with these \textit{soft} linear interpolations. Formally, the objective can be written as Eq. \ref{eq:mupbm_loss}, where $q(y_{si} |x, x' )$ denotes distribution defined by interpolated softmax $y_{si}=\beta y + (1 - \beta) y'$.
\begin{equation}
    \mathcal{L}_U (S, T; \phi, \psi) = \mathbb{E}_{(x, x') \sim \mathcal{P}_{S, T}, t \sim \mathcal{T}_{si}} [D_{KL}(p_{\theta}(y\ | \ t(x, x', \beta)) \ || \ q(y_{si} |x, x' ))]
    \label{eq:mupbm_loss}
\end{equation}
\subsection{Other Predictive Behavior Matching}
\label{sub:Auxiliary Semantic Information Matching}
There has been some other works align with our intuition of PB matching~\cite{sun2019unsupervised, carlucci2019domain}. For example, introduced by~\cite{sun2019unsupervised}, self-supervised tasks also describe a predictive behavior that should be identical between domains, which we call as Task-oriented Predictive Behavior Matching hereafter. Following \cite{sun2019unsupervised}, we apply three semantic transforming operations (\textit{Rotation Prediction}, \textit{Vertical Flip Prediction}, \textit{Patch Location Prediction}) to realize this objective. We specify the operations as $t \sim \mathcal{T}_{st}$ and its self-supervised label $\tilde{y}_t = t(x_{S, T})$. We denote each corresponding auxiliary head as $h_t$ parameterized by $\omega$ for each $t(x_{S, T})$, and obtain the following training objective with supervised loss function $L_t$:
\begin{equation}
    \mathcal{L}_S (S, T; \phi, \omega) = \mathbb{E}_{x \sim \mathcal{P}_{S, T}, t \sim \mathcal{T}_{st}} [L_t(h_t \circ g(t(x_{S, T})), \tilde{y}_t)]
    \label{self-training-loss}
\end{equation}

To summarize, InstaPBM investigated the potential of PB matching for robust DA. Additionally, it has the potential to add more PB matchings to further enhance robustness. The overall optimization is defined as in Eq.~\ref{eq:total_objectives}, where $\omega$ is the parameters of predictor $h_t$.
\begin{equation}
    \mathop{\text{min}} \limits_{\theta, \phi, \omega} \lambda_M \mathcal{L}_M (S, T; \theta) + \lambda_C \mathcal{L}_C (S, T; \theta)  + \lambda_U \mathcal{L}_U(S, T; \phi, \psi) + \lambda_S \mathcal{L}_S (S, T; \phi, \omega)
    \label{eq:total_objectives}
\end{equation}

\section{Experiments}
\label{section:Experiments}

We evaluate the limitations of DM based methods and the efficacy of InstaPBM on both conventional and RDS benchmarks. More experimental details are included in the supplementary material.

\subsection{Experimental Settings}
\noindent\textbf{Conventional Benchmarks.} We experiment on two common conventional DA benchmarks: Digits5~\cite{lecun1998gradient, ganin2015unsupervised, arbelaez2011contour, netzer2011reading} and  VisDA2017~\cite{peng2017visda}. Digits5 consists of five vastly different domains: MNIST (MN), MNISTM (MM), USPS (US), SYNTH (SY), SVHN (SV). 
Each domain contains more than 50K images except USPS, which has about 10K images. VisDA2017 is a challenging testbed for UDA with domain shift from simulation to real, which has around 280K images from 12 classes. 

\noindent\textbf{Proposed RDS Benchmarks.} We propose the following RDS benchmarks: (1) \textbf{Label Distribution Shift:} Digits5-LDS, VisDA2017-LDS, and DomainNet-LDS by respectively resampling Digits5, VisDA2017, and DomainNet~\cite{peng2019moment} under LDS protocol; (2) \textbf{Intermediate layer Distribution Shift:} DomainNet-ILDS by combining similar labels into a parental label on DomainNet; C, R, S, P, I, Q represent six subdomains: Clipart, Real, Sketch, Painting, Infograph and Quickdraw respectively. (3) \textbf{Target with Outliers:} ID-TwO by selecting 18 overlapped classes between the source ImageNet~1K and target DomainNet for close-set UDA. Due to space limitation, statistics of RDS benchmarks and results on VisDA2017-LDS and Digits5-LDS will be provided in supplementary material.

\noindent\textbf{Baselines.} We compare with the following state-of-the-art methods: DM methods: DANN~\cite{revgrad}, CoRAL~\cite{sun2016return, dcoral, morerio2018minimal}, DAN~\cite{long2015learning}, HoMM~\cite{chen2019homm}; conditional DM: CDAN\cite{long2018conditional}, sDANN-$\beta$~\cite{wu2019domain}, JAN~\cite{long2017deep}; none DM:   ADR~\cite{saito2017adversarial}, SE~\cite{french2017self} and CAN~\cite{kang2019contrastive} on aforementioned benchmarks.




\noindent\textbf{Implementation Details.} We use LeNet, ResNet-101, and ResNet-101 as backbone networks for DA tasks on Digits5, VisDA2017, and DomainNet, respectively. We replace the last fully-connected (FC) layer with the task-specific FC layer. Please find more details in the supplementary material.


\subsection{Results and Analysis}

\begin{table}[!t]
\renewcommand\tabcolsep{1.5pt} 
\centering\scriptsize 
\caption{Performance comparison on \textbf{Digits5} dataset. Best accuracy in bold.}
\resizebox{\linewidth}{!}{%
\begin{tabular}{c |cccc|cccc|cccc|cccc|cccc| c }
\toprule
 & \multicolumn{4}{c|}{\centering MN} & \multicolumn{4}{c|}{\centering MM} & \multicolumn{4}{c|}{\centering US} & \multicolumn{4}{c|}{\centering SY} & \multicolumn{4}{c|}{\centering SV} \\
\cline{2-21}
Method &  MM &  US &  SY &  SV &  MN &  US &  SY &  SV &  MN &  MM &  SY &  SV &  MN &  MM &  US &  SV &  MN &  MM &  US &  SY & Average \\
\hline
Source Only & 67.7 & 73.9 & 54.2 & 47.3 & 81.8 & 73.4 & 50.9 & 50.7 & 75.6 & 50.9 & 42.5 & 41.6 & 85.2 & 65.0 & 82.1 & 75.6 & 56.8 & 44.8 & 63.2 & 73.2 & 62.8 \\
\hline 
DAN~\cite{long2015learning} & 62.6 & 79.7 & 47.6 & 43.1 & 99.2 & 80.5 & 60.5 & 53.3 & 48.2 & 28.3 & 28.2 & 24.4 & 88.6 & 65.8 & 86.6 & 79.3 & 74.5 & 56.2 & 70.1 & 93.3 & 63.5 \\
DANN~\cite{revgrad} & 64.9 & 75.0 & 52.3 & 42.5 & 89.0 & 71.3 & 49.7 & 58.6 & 67.1 & 53.7 & 45.3 & \textbf{40.1} & 81.3 & 63.6 & 79.0 & 83.2 & 70.2 & 55.5 & 63.2 & 77.5 & 64.2 \\
CoRAL~\cite{dcoral} & 70.2 & 83.9 & 54.2 & 45.8 & 98.7 & 78.4 & 59.5 & 53.0 & 72.6 & 51.3 & 41.9 & 33.8 & 87.5 & 67.4 & 82.8 & 84.5 & 68.5 & 53.2 & 67.4 & 90.8 & 67.3 \\
HoMM~\cite{chen2019homm} & 72.2 & 86.2 & 58.0 & 50.5 & 98.2 & 79.6 & 58.7 & 51.6 & 75.2 & 49.1 & 43.6 & 37.7 & 88.3 & 68.4 & 83.5 & 82.4 & 68.7 & 60.4 & 65.5 & 88.8 & 68.3 \\
\hline 
sDANN-$\beta$~\cite{wu2019domain} & 50.2 & 70.4 & 44.5 & 38.3 & 79.0 & 70.5 & 52.9 & 53.4 & 74.6 & 48.9 & 34.5 & 36.6 & 83.4 & 60.6 & 79.0 & 71.6 & 60.2 & 47.8 & 42.8 & 64.5 & 58.2 \\
JAN~\cite{long2017deep} & 66.2 & 88.5 & 47.9 & 47.7 & 98.4 & 80.8 & 59.6 & 54.2 & 68.8 & 32.6 & 38.1 & 32.2 & 87.2 & 57.7 & 83.0 & 79.9 & 71.5 & 52.9 & 69.8 & 90.8 & 65.4 \\
CDAN~\cite{long2018conditional} & 75.7 & 85.5 & \textbf{65.5} & \textbf{54.0} & 99.0 & 79.7 & 62.6 & 56.1 & 69.0 & 51.3 & \textbf{46.3} & 38.4 & 89.6 & 71.8 & 84.6 & 83.4 & 73.8 & 58.2 & 72.0 & 91.4 & 70.4 \\
CDAN+E~\cite{long2018conditional}  & 70.7 & 90.8 & 63.4 & 49.3 & 99.5 & 88.5 & 73.1 & 63.5 & 87.0 & 46.8 & 46.2 & 34.0 & 86.0 & 65.4 & 67.0 & 80.5 & 81.5 & 62.7 & 71.8 & 95.9 & 71.2 \\
CAN~\cite{kang2019contrastive} & 82.4 & 94.6 & 45.8 & 42.4 & 99.5 & 70.7 & 71.2 & 61.8 & 88.1 & 53.3 & 39.8 & 31.1 & 94.6 & 81.8 & 94.1 & 82.3 & 97.1 & 89.4 & 73.0 & 95.3 & 74.4 \\
\hline 
ADR~\cite{saito2017adversarial} & 83.6 & 93.4 & 58.8 & 50.1 & \textbf{99.6} & \textbf{84.9} & 72.6 & 37.3 & 48.6 & 43.4 & 38.6 & 37.6 & 97.0 & 84.1 & 94.6 & 81.9 & 83.4 & 88.1 & 73.8 & \textbf{97.6} & 72.5 \\
SE~\cite{french2017self} & 81.7 & 94.2 & 38.7 & 34.7 & 99.4 & 85.4 & 90.1 & 60.1 & 88.2 & 53.6 & 36.4 & 30.0 & 95.0 & 83.2 & 91.3 & 80.7 & 77.2 & 87.2 & 67.4 & 95.4 & 73.5 \\

\textbf{InstaPBM(ours)} & \textbf{94.6} & \textbf{96.5} & 45.7 & 45.9 & 99.2 & 67.3 & \textbf{91.1} & \textbf{77.7} & \textbf{92.0} & \textbf{58.6} & 42.0 & 32.3 & \textbf{97.2} & \textbf{89.2} & \textbf{94.8} & \textbf{87.0} & \textbf{97.6} & \textbf{91.7} & \textbf{80.0} & 96.6 & \textbf{78.9} \\
\hline
\end{tabular}
}
\label{tab:LENet_digits20}
\end{table}
\begin{table}[!t]
\centering\footnotesize
\caption{Performance comparison on \textbf{VisDA2017} dataset. Best accuracy in bold.}
\resizebox{\linewidth}{!}{%
\begin{tabular}{c |cccccccccccc| c }
\toprule
 Method &  {\scriptsize a-plane} &  {\scriptsize bicycle} &  {\scriptsize bus} &  {\scriptsize car} &  {\scriptsize horse} &  {\scriptsize knife} &  {\scriptsize m-cycle} &  {\scriptsize person} &  {\scriptsize plant} &  {\scriptsize skb} &  {\scriptsize train} &  {\scriptsize truck} & Average \\
\hline
 Source Only & 72.3 & 6.1 & 63.4 & 91.7 & 52.7 & 7.9 & 80.1 & 5.6 & 90.1 & 18.5 & 78.1 & 25.9 & 49.4 \\
 \hline
 DANN~\cite{revgrad} & 81.9  & 77.7  & 82.8  & 44.3  & 81.2  & 29.5  & 65.1  & 28.6  & 51.9  & 54.6  & 82.8  & 7.8  & 57.4  \\
 DAN~\cite{long2015learning} & 68.1  & 15.4  & 76.5  & 87.0  & 71.1  & 48.9  & 82.3  & 51.5  & 88.7  & 33.2  & 88.9  & 42.2  & 62.8  \\
 JAN~\cite{long2017deep} & 75.7  & 18.7  & 82.3  & 86.3  & 70.2  & 56.9  & 80.5  & 53.8  & 92.5  & 32.2  & 84.5  & 54.5  & 65.7  \\
 MCD\cite{Saito_2018_CVPR} & 87.0  & 60.9  & 83.7  & 64.0  & 88.9  & 79.6  & 84.7  & 76.9  & 88.6  & 40.3  & 83.0  & 25.8  & 71.9  \\
 ADR~\cite{saito2017adversarial} & 87.8  & 79.5  & 83.7  & 65.3  & 92.3  & 61.8  & 88.9  & 73.2  & 87.8  & 60.0  & 85.5  & 32.3  & 74.8  \\
 SE~\cite{french2017self} & 95.9  & \textbf{87.4}  & 85.2  & 58.6  & 96.2  & 95.7  & 90.6  & 80.0  & 94.8  & 90.8  & 88.4  & 47.9  & 84.3 \\
 CAN~\cite{kang2019contrastive} & \textbf{97.0}  & 87.2  & 82.5  & 74.3  & \textbf{97.8}  & \textbf{96.2}  & 90.8  & 80.7  & \textbf{96.6}  & \textbf{96.3}  & 87.5  & 59.9  & 87.2  \\
 \textbf{InstaPBM(ours)} & 95.4 & 86.3 & \textbf{94.0} & \textbf{93.1} & 92.8 & 95.6 & \textbf{92.2} & \textbf{87.5} & 92.8 & 85.5 & \textbf{92.4} & \textbf{85.0} & \textbf{91.1} \\
\hline
\end{tabular}
}
\label{tab:RESNET101_VisDA2017}
\end{table}


\subsubsection{Conventional Benchmarks}
Table~\ref{tab:LENet_digits20} and Table~\ref{tab:RESNET101_VisDA2017} show the performance comparison on Digits5 and VisDA2017 respectively. From the results, we observe that: (1) The source-only method that directly transfers the model trained on source to target performs the worst. This is reasonable, because it does not address the domain shift. (2) On DIGITS, InstaPBM achieves 4.5\% performance gain as compared to the best baseline (\textit{i.e.} CAN). Please note that many UDA methods~\cite{dirtt2018iclr, sun2019unsupervised} only experiment on several selected scenarios (\textit{e.g.} MNIST->SVHN, USPS->MNIST), which cannot sufficiently verify their efficacy and generalization. (3) On VisDA2017, InstaPBM obtains the best accuracy on average: 3.9\% and 6.8\% absolute improvements compared to the best baseline CAN~\cite{kang2019contrastive} and the self-embedding (SE)~\cite{french2017self}, which wins the first place in VisDA-2017 leaderboard. Please note that we do not use any engineering trick (e.g. ensembling, parameters grid searching, warm-up, etc).




\subsubsection{Realistic Domain Shift (RDS) Benchmarks}
\label{subsub: realistic_domain_shift}
\noindent\textbf{Label Distribution Shift (LDS).} From the left part results in Table~\ref{tab:RESNET50_domainnet_LDS_ILDS}, we observe that: (1) Almost all DM methods have the detrimental effects on adaptation under LDS, which verifies their limitations analyzed in Sec.~\ref{section:Limitations_of_DM}. (2) Conditional DM methods perform relatively better than DM methods, which demonstrates they are more robust to LDS. (3) None DM methods are least affected by LDS and InstaPBM performs the best, which indicates its strong resistance towards LDS.

\noindent\textbf{Intermediate Layer Distribution Shift (ILDS).} The right part in Table~\ref{tab:RESNET50_domainnet_LDS_ILDS} shows the performance comparison on DomainNet-ILDS. Similar to LDS scenarios, the DM methods and none DM methods have a weak and strong resistance towards ILDS respectively. What differs from the results of LDS is that conditional DM methods also result in adaptation performance decay on ILDS Benchmarks. Our method still achieves the best result on ILDS Benchmarks as compared to all other baselines.

\noindent\textbf{Target with Outliers (TwO).} The result in Table~\ref{tab:RESNET101_realistic} verifies our analysis that when target data are filled with outliers from other domains, DM will suffer interventions. None DM methods, such as our InstaPBM, are less affected since they do not rely on matching representation distributions. For outliers, we can better capture their semantics by contrastively matching predictive behaviors with source domain data per instance.
\begin{table}[!t]
\centering\footnotesize
\caption{Performance comparison on \textbf{DomainNet-LDS} and \textbf{DomainNet-ILDS} Benchmarks.}
\resizebox{\linewidth}{!}{%
\begin{tabular}{c |cccccc|c||cccccc|c }
\toprule
& \multicolumn{7}{c||}{\centering LDS Benchmark} & \multicolumn{7}{c}{\centering ILDS Benchmark} \\
\cline{2-15}
 Method & I->C & C->R & C->S & S->P & P->R & R->C & Average & I->C & C->R & C->S & S->P & P->R & R->C & Average \\
\hline
Source Only & 31.2 & 52.5 & 46.3 & 39.9 & 53.7 & 49.0 & 45.4 & 40.9 & 62.3 & 58.9 & 49.0 & 64.7 & 60.4 & 56.0 \\
\hline
DAN~\cite{long2015learning} & 22.0 & 46.3 & 42.2 & 36.6 & 47.6 & 46.6 & 40.2 & 32.3 & 57.6 & 52.0 & 46.7 & 61.8 & 56.5 & 51.1 \\
CoRAL~\cite{dcoral} & 22.5 & 47.1 & 44.6 & 38.3 & 49.3 & 46.3 & 41.3 & 35.2 & 60.1 & 53.8 & 49.2 & 63.0 & 59.1 & 53.4 \\
HoMM~\cite{chen2019homm} & 21.5 & 47.8 & 45.2 & 38.5 & 48.4 & 47.7 & 41.5 & 37.2 & 61.2 & 55.9 & 50.2 & 58.6 & 61.3 & 54.1 \\
DANN~\cite{revgrad} & 27.2 & 51.6 & 46.6 & 40.1 & 51.4 & 46.3 & 43.9 & 40.4 & 62.2 & 54.6 & 51.5 & 62.6 & 59.5 & 55.1 \\
\hline
JAN~\cite{long2017deep} & 24.7 & 50.3 & 43.1 & 37.4 & 49.3 & 48.2 & 42.2 & 38.4 & 60.5 & 55.4 & 46.6 & 58.6 & 59.9 & 53.2 \\
CDAN~\cite{long2018conditional} & 30.0 & 53.2 & 46.9 & 41.4 & 54.3 & 48.0 & 45.6  & 37.4 & 59.5 & 53.2 & 46.7 & 61.3 & 56.5 & 52.4 \\
sDANN-$\beta$~\cite{wu2019domain} & 32.0 & 52.1 & 47.2 & 45.2 & 50.3 & 49.7 & 46.1 & 39.1 & 60.2 & 53.5 & 47.2 & 60.4 & 60.1 & 54.2 \\
CDAN+E~\cite{long2018conditional} & 29.7 & 55.3 & 48.1 & 42.4 & 54.3 & 53.2 & 47.2 & 38.6 & 61.6 & 53.9 & 45.9 & 62.8 & 58.6 & 53.6 \\
CAN~\cite{kang2019contrastive} & 36.0 & 58.2 & \textbf{53.2} & \textbf{47.3} & 53.6 & 57.6 & 51.0 & 40.1 & 64.6 & 55.4 & 47.3 & 63.2 & 59.1 & 56.5 \\
\hline
ADR~\cite{saito2017adversarial} & 34.5 & 54.6 & 47.7 & 42.3 & 56.6 & 51.3 & 47.8  & 43.2  & 65.4  & 57.1  & 51.4  & 66.8  & 58.0  & 57.0 \\
SE~\cite{french2017self} & 33.6 & 56.6 & 50.3 & 40.6 & 55.5 & 55.1 & 48.6  & 45.3 & 67.2 & 61.3 & 52.4 & 70.0 & 66.3 & 60.4 \\
\textbf{InstaPBM(ours)} & \textbf{38.5} & \textbf{60.2} & 52.3 & 47.0 & \textbf{61.6} & \textbf{59.6} & \textbf{53.2}  & \textbf{49.6} & \textbf{72.1} & \textbf{64.1} & \textbf{53.1} & \textbf{72.3} & \textbf{68.7} & \textbf{63.3} \\
\hline
\end{tabular}
}
\label{tab:RESNET50_domainnet_LDS_ILDS}
\end{table}
\begin{table}[!t]
\renewcommand\tabcolsep{2 pt} 
\centering\footnotesize
\caption{Performance comparison on \textbf{ID-TwO} based on ResNet-101. Best accuracy in bold.}
\resizebox{\linewidth}{!}{%
\begin{tabular}{c|cccccccccccccccccc|cc}
\toprule
 Method & {\scriptsize{a-plane}} & {\scriptsize bicycle} & {\scriptsize clock} & {\scriptsize{dog}} & {\scriptsize f-pan} & {\scriptsize lion} & {\scriptsize n-lace} & {\scriptsize potato} & {\scriptsize sock} & {\scriptsize b-ball} & {\scriptsize bucket} & {\scriptsize cup} & {\scriptsize d-bell} & {\scriptsize laptop} & {\scriptsize lipstick} & {\scriptsize panda} & {\scriptsize shark} & {\scriptsize w-bottle} & Average \\
\hline
Source Only & 52.8 & 69.5 & 67.9 & 61.2 & 51.6 & 56.3 & 94.0 & 30.2 & 53.7 & 46.8 & 57.5 & 59.9 & 61.3 & 57.3 & 55.0 & 47.8 & 45.7 & 59.8 & 57.1 \\
\hline 
DAN~\cite{long2015learning} & 49.3 & 61.5 & 68.1 & 63.0 & 56.9 & 51.1 & 83.8 & 37.6 & 51.3 & 42.1 & 53.3 & 60.9 & 57.2 & 62.1 & 51.8 & 42.0 & 41.6 & 57.3 & 55.1 \\
DANN~\cite{revgrad} & 56.7 & 82.0 & 76.3 & 62.7 & 51.3 & 54.4 & 88.1 & 35.7 & 55.7 & 43.7 & 72.1 & 57.0 & 63.3 & 57.5 & 54.6 & 43.6 & 44.4 & 59.8 & 58.8 \\
HoMM~\cite{chen2019homm} & 54.0 & 75.9 & 68.1 & 59.8 & 53.2 & 57.9 & 93.1 & 36.2 & 55.4 & 54.1 & 73.1 & 58.4 & 68.1 & 67.7 & 54.8 & 46.5 & 57.2 & 62.1 & 60.9 \\
CoRAL~\cite{dcoral} & 53.7 & 91.4 & 89.3 & 70.6 & 62.4 & 53.7 & 69.8 & 42.9 & 57.3 & 40.0 & 63.3 & 58.1 & 62.3 & 60.0 & 58.0 & 52.4 & 54.1 & 61.4 & 61.1 \\
\hline
sDANN-$\beta$~\cite{wu2019domain} & 58.4 & 72.7 & 78.4 & 42.6 & 64.2 & 50.1 & 81.4 & 38.4 & 50.5 & 51.0 & 69.4 & 48.1 & 46.4 & 56.1 & 59.5 & 49.2 & 49.5 & 76.0 & 57.8 \\
JAN~\cite{long2017deep} & 53.7 & 69.7 & 76.5 & 62.6 & 67.2 & 55.1 & 85.2 & 40.4 & 56.5 & 51.1 & 71.9 & 50.1 & 64.4 & 65.5 & 56.5 & 44.7 & 47.5 & 68.0 & 60.4 \\
CDAN~\cite{long2018conditional} & 55.5 & 87.0 & 73.2 & 62.5 & 61.6 & 54.9 & 83.3 & \textbf{59.6} & 58.6 & 53.5 & 74.2 & 70.0 & 69.3 & 69.2 & 56.8 & 44.6 & \textbf{77.4} & 74.0 & 65.8 \\
CDAN+E~\cite{long2018conditional} & 55.8 & 76.5 & 77.4 & 70.7 & 61.6 & 58.4 & 89.6 & 47.6 & 61.4 & 61.1 & 75.4 & 67.1 & 81.9 & 71.1 & 55.3 & 46.1 & 52.9 & 78.6 & 66.0 \\
CAN~\cite{kang2019contrastive} & 60.7 & 93.2 & 88.8 & \textbf{74.5} & 77.3 & 59.8 & 87.7 & 52.1 & 79.3 & 84.2 & 83.2 & 72.0 & 81.4 & 70.6 & 58.2 & 59.5 & 62.9 & 83.5 & 73.8 \\
\hline 
ADR~\cite{saito2017adversarial} & 57.7 & 85.8 & 82.1 & 73.5 & 73.0 & 62.7 & 87.5 & 52.6 & 73.6 & 68.2 & 79.4 & 67.5 & 79.6 & 77.3 & 53.1 & 55.1 & 52.1 & 77.3 & 69.9 \\
SE~\cite{french2017self} & 58.3 & 83.9 & 84.9 & 71.3 & 73.5 & 65.0 & 86.7 & 53.1 & 77.6 & 74.0 & 81.5 & \textbf{71.4} & 77.8 & 74.3 & 56.3 & 57.7 & 55.3 & 81.4 & 71.3 \\
\textbf{InstaPBM(ours)} & \textbf{62.7} & \textbf{96.3} & \textbf{91.3} & 72.7 & \textbf{76.3} & \textbf{66.3} & \textbf{91.5} & 55.4 & \textbf{86.2} & \textbf{86.8} & \textbf{87.5} & 67.0 & \textbf{84.9} & \textbf{92.7} & \textbf{61.8} & \textbf{61.8} & 59.4 & \textbf{92.2} & \textbf{77.4} \\
\hline
\end{tabular}
}
\label{tab:RESNET101_realistic}
\end{table}
\begin{table}[!t]
\centering\footnotesize
\caption{Ablation study on different components in InstaPBM across all benchmarks.}
\resizebox{\linewidth}{!}{%
\begin{tabular}{c |cc|ccccc| c }
\toprule
& \multicolumn{2}{c|}{Conventional Benchmarks} & \multicolumn{5}{c|}{RDS Benchmarks} \\
\cline{2-8}
Method & VisDA2017 & Digits5 & V-LDS & D-LDS & DN-LDS & DN-ILDS & ID-TwO & Average \\
\hline
Baseline & 49.4 & 62.8 & 55.1 & 55.8 & 45.4 & 56.0 & 57.1 & 54.5 \\
\hline
+MIM & 83.5 & 72.1 & 69.8 & 64.3 & 52.9 & 61.1 & 73.8 & 68.2 \\
\hline
+CPBM\_RA & 77.9 & 67.5 & 62.2 & 59.5 & 50.7 & 57.7 & 69.5 & 63.6 \\
+CPBM\_NI & 71.2 & 65.7 & 63.3 & 57.1 &  48.1 & 56.7 & 65.1 & 61.0 \\
+CPBM\_ALL & 79.2 & 70.2 & 64.5 & 60.4 & 50.9 & 59.0 & 69.9 & 64.9 \\
\hline
+MuPBM & 75.3 & 67.1 & 59.5 & 58.9 & 49.2 & 58.7 & 64.3 & 61.8 \\
\hline
+TPBM\_ROT & 62.3 & 64.3 & 56.1 & 57.2 & 45.7 & 56.1 & 60.5 & 57.5 \\
+TPBM\_QDR & 64.6 & 66.2 & 58.2 & 59.3 & 46.1 & 56.7 & 61.3 & 58.9 \\
+TPBM\_FLIP & 59.8 & 62.6 & 55.9 & 54.3 & 45.5 & 56.2 & 58.3 & 56.1 \\
+TPBM\_ALL & 65.5 & 67.3 & 58.5 & 60.0 & 46.5 & 57.7 & 61.8 & 59.6 \\
\hline
+InstaPBM & 91.1 & 78.9 & 74.7 & 66.2 & 53.2 & 63.3 & 77.4 & 72.1 \\
\hline
\end{tabular}
}
\label{tab:RESNET101_contributive_parts}
\tiny{\textbf{Baseline} denotes model trained on source without adaptation. \textbf{+MIM} denotes using our mutual information maximization objective. \textbf{+MuPBM} represents Mix-up Predictive Behavior Matching. \textbf{+CPBM\_*} denotes applying Contrastive Predictive Behavior Matching by Random Augmentation(RA), Noise Injection(NI) and both(ALL) specifically. \textbf{+TPBM\_*} denotes implementing Task-oriented Predictive Behavior Matching by using different self-supervised tasks(i.s. Rotation Prediction(ROT), Vertical Flip Prediction(FLIP) and Patch Location Prediction(QDR)). \textbf{V-LDS} denotes \textbf{VisDA2017-LDS}, \textbf{D-LDS} denotes \textbf{Digits5-LDS}, \textbf{DN-LDS} and \textbf{DN-ILDS} denote \textbf{DomainNet-LDS} and \textbf{DomainNet-ILDS} respectively. \textbf{InstaPBM} denotes the integration of MIM, MuPBM, CPBM\_ALL and TPBM\_ALL.
}
\end{table}

\subsection{Ablation Study}

Our approach has four main components: MIM, CPBM, MuPBM and TPBM. We conduct an exhaustive ablation study to explore the specific contribution of each component. The results on 7 different benchmarks are shown in Table~\ref{tab:RESNET101_contributive_parts}. We have the following observations: (1) Comparing to Source Only method, each proposed component(From 2 to 10 rows) can improve the classification accuracy on all benchmarks, which demonstrates that these components can not only improve the domain adaptation performance but also have a strong resistance towards RDS; (2) Among all these components, MIM obtains the highest performance improvement (13.7\%) followed by CPBM\_ALL (10.4\%); (3) MIM, CPBM, MuPBM and TPBM are not mutually exclusive. When jointly employing all of them(The final row), we can obtain the the best performance on all benchmarks.


\section{Conclusions}
In this work, we systematically analyze the limitations of DM algorithms under more complex and realistic domain shifts. We argue that the most popular method for domain adaptation - Distribuitonal Matching relies on restrictive assumptions on underlying distributions and are usually not satisfied in real world. Thus we propose to replace matching distributions by matching discriminative patterns of predictors (PBs). 

For quantitatively analysis, we propose new RDS benchmarks and provide extensive empirical evaluations. To overcome these limitations and promote robust DA under RDS, we propose InstaPBM to match predictive behaviors across domains. Extensive experiments on both conventional and RDS benchmarks demonstrate the efficacy of InstaPBM. We hope our intuitive observations on DM methods and effective methods proposed in InstaPBM will serve as a new direction and achieve robust DA in real scenarios.

\section{Broader Impact}

Supervised learning algorithms suffer from low accuracy when the target test data is out of distribution of the source train data or lack of annotations. Domain adaptation (DA) is an actively researched area to solve this problem by leveraging the unlabeled target data and adapt the model to produce powerful predictions. In this work, we systematically analyze the limitations of the traditional distributional matching based design pattern for domain adaptation, discuss its flaws, and find it usually fails under real world domain shift. Such analysis will be helpful to avoid failures brought by the distributional matching based methods. 
To quantitatively verify the limitations of the distributional matching based methods, we propose new DA benchmarks with three types of Realistic Domain Shifts (RDS), which better reflect the real world scenarios and can promote DA research with more practical effectiveness.
We propose a more intuitive and effective domain adaptation approach, which is closer to the essential human behavior and more robust to the real world scenarios. 
By achieving robust domain adaptation, our method can help machine learning model transfer the learned knowledge to new scenes. Since most distributional matching based DA methods suffer from instability in training and delicacy of special parameters, our proposed DA pattern are more robust, which is unquestionably more applicable in real industrial deployment. As for negative aspects, since the method is designed for unsupervised domain adaptation, it may not be able to achieve significant performance gain when there is several labeled samples in the target domain.
\bibliographystyle{unsrtnat}
\bibliography{refs}
\end{document}